\documentclass[runningheads]{llncs}

\usepackage[colorlinks]{hyperref}
\usepackage{url}
\usepackage{amsmath}
\usepackage{algorithm}
\usepackage{algorithmic}

\usepackage{lipsum}
\usepackage{multirow}
\usepackage{booktabs}
\usepackage{graphicx}
\usepackage[normalem]{ulem}
\usepackage{amsmath}
\usepackage{amssymb}
\usepackage{mathrsfs}
\usepackage{float}
\usepackage[colorinlistoftodos]{todonotes}

\usepackage[misc]{ifsym}

\newcommand{\name}{DPMN}
\begin{document}

\title{Deep Prompt Multi-task Network for Abuse Language Detection}
\titlerunning{\name}
%
\author{Jian Zhu \and
Yuping Ruan \and
Jingfei Chang \and 
Wenhui Sun \and
Hui Wan \and
Jian Long \and
Cheng Luo\inst{*}}
\authorrunning{F. Author et al.}
%
\institute{
Zhejiang Lab \\ \email{\{qijian.zhu, ypruan, cjf\_chang, sunwh, wanhui, longjian, luo\_cheng\}@zhejianglab.com} }
\toctitle{DPMN}
\tocauthor{}
\authorrunning{Jian Zhu et al.}

%

%
%

%

\maketitle

\def\thefootnote{*}\footnotetext{Corresponding author.}
\begin{abstract}
The detection of abuse language remains a long-standing challenge with the extensive use of social networks. The detection task of abuse language suffers from limited accuracy. We argue that the existing detection methods utilize the fine-tuning technique of the pre-trained language models (PLMs) to handle downstream tasks. Hence, these methods fail to stimulate the general knowledge of the PLMs. To address the problem, we propose a novel \underline{D}eep \underline{P}rompt \underline{M}ulti-task \underline{N}etwork (DPMN) for abuse language detection. Specifically, DPMN first attempts to design two forms of deep prompt tuning and light prompt tuning for the PLMs. The effects of different prompt lengths, tuning strategies, and prompt initialization methods on detecting abuse language are studied. In addition, we propose a Task Head based on Bi-LSTM and FFN, which can be used as a short text classifier. Eventually, DPMN utilizes multi-task learning to improve detection metrics further. The multi-task network has the function of transferring effective knowledge. The proposed DPMN is evaluated against eight typical methods on three public datasets: OLID, SOLID, and AbuseAnalyzer. The experimental results show that our DPMN outperforms the state-of-the-art methods.
\end{abstract}
\section{Introduction}

\subsection{Background}
The abuse language has spread throughout social media and become a significant issue. On social network sites like Facebook, Twitter, and Instagram, some groups become targets of online bullying activities. Any expression that denigrates or offends a person or group of people is referred to as abuse, and examples include sexism, harassment, cyberbullying, personal insults, racism, etc. Abuse language can be directed at particular people or groups. Abuse language can have serious psychological consequences for the victim and hinder freedom of expression. Intelligent detection algorithms can identify abuse content in a significant volume of social media. It is essential to minimize the psychological toll on victims to stop hate crimes. As a result, it is important to intelligently detect and govern abuse language before it spreads on social networks. Past research has examined various abuse language issues, including abuse and hate speech.

Abuse language detection can be seen as short text classification. As natural language processing evolves, detecting abuse language can be roughly divided into three periods. Early detection algorithms adopt conventional machine learning methods, and the performance of model detection results largely depends on the features of manual design. These hand-designed features mainly include character features, word features, n-gram features, syntactic features, and linguistic features. The second phase of the detection algorithm uses the deep learning method. Typical deep networks are CNN \cite{lecun:62}, and RNN \cite{mikolov:51}. The advantage of the deep learning method is that it does not need to design features manually. It can automatically generate context features of short text through the deep neural network. Third, the large PLMs, such as BERT \cite{devlin:22} and GPT \cite{radford:57}, improve the metric of detecting abuse language. Because these models are trained on a large-scale corpus, general knowledge of natural language can be obtained.

Due to the complicated of natural language laws, automatically identifying abuse language is still exceedingly challenging. For example, abuse language generally occurs in two cases, explicit and indirect linguistic phenomena. The earlier type of abuse language is more overt, perhaps taking the form of specific harsh phrases. However, the latter type could contain metaphors or analogies, which might cause certain algorithmic identification mistakes. In addition, the existing methods focus on fine-tuning the PLMs to adapt to the downstream tasks so that the training and application of the PLMs are not under a unified paradigm, and the knowledge contained in the PLMs cannot be better utilized. Therefore, the primary goal of our work is to do this research: whether prompt-based learning helps detect abuse language. To this end, we propose a new end-to-end multi-task detection network for abuse language, which combines prompt tuning and multi-task learning.

\subsection{Motivations and Contributions}
The task of abuse language detection suffers from limited accuracy. Current detection methods \cite{liu:26,chandra:17,dai:17,wiedemann:55,hakimov:54} fine-tune the PLMs to adapt to downstream tasks. NULI \cite{liu:26} adapts and fine-tunes the BERT-base model to detect abuse language. AbuseAnalyzer \cite{chandra:17} uses a two-layer feed-forward network with BERT for detecting abuse language. Kungfupanda \cite{dai:17} develops a method for detecting abuse language that blends multi-task learning with BERT-based models. With regard to their effectiveness in detecting abuse language, UHH-LT \cite{wiedemann:55} uses the MLM method to compare the performance of different PLMs. An architecture called CTF \cite{hakimov:54} combines various textual elements to find abuse or hostile tweets on Twitter, which generates contextual 768-dimensional word vectors for each input character using a pre-trained BERT model.
Therefore, these methods fail to stimulate the general knowledge of the PLMs, leading to limited accuracy. To address the mentioned problem, we introduce prompt-based learning and explore how prompt-based learning can be correctly used in detecting abuse language. 

\begin{figure}[htp]
  \centering
  \includegraphics[width=7cm]{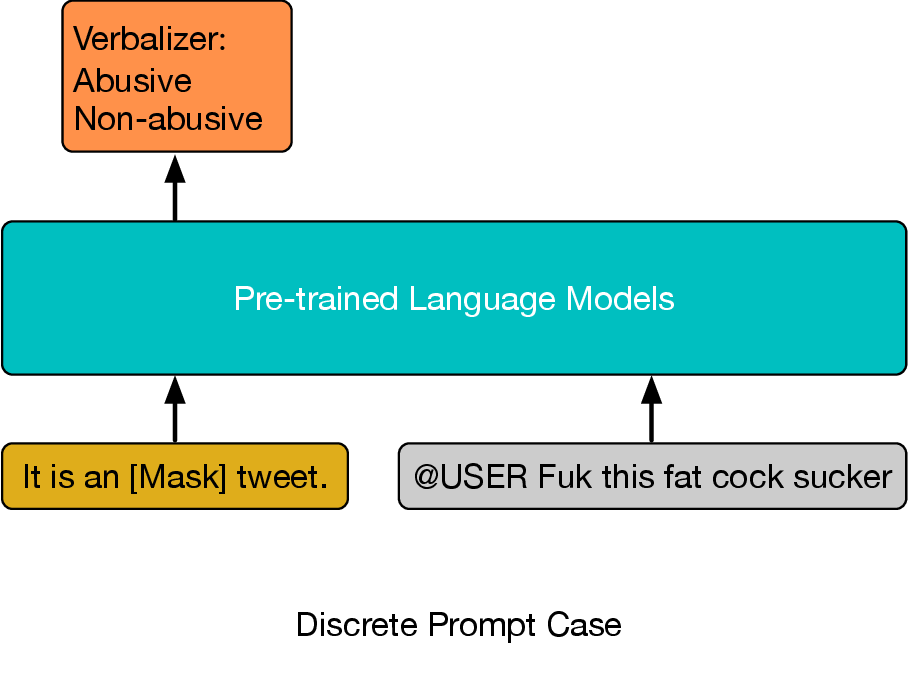}
  \caption{An example of applying prompt-based learning to identify whether a tweet is an abuse language. The mask character is a word to be predicted. It can have two-word choices in the verbalizer, abuse and non-abuse.}
  \label{fig:00}
\end{figure}

Prompt tuning has been a great success for most natural language processing tasks. By including new texts in the input, prompt-based learning is a method for better using the knowledge from the PLMs. As shown in Fig. \ref{fig:00}, a prompt with a mask token is added to the tweet text, and we predict that the mask would point to the corresponding word in the vocabulary through PLMs. Then, based on the corresponding label of the word, it can determine whether the tweet is abuse language. 

We propose a novel \textit{Deep Prompt Multi-task Network} termed DPMN. First, it uses prompt-based learning to acquire knowledge of the PLMs. Second, to use the supervisory signals from other related tasks, we employ multi-task learning.
Eventually, we design a task head based on the synthesis of Bi-LSTM \cite{zhang:52}, and feed-forward network (FFN) \cite{bebis:53} to aggregate all the shared representations of the final output layer of the BERT model. 

Experimental results indicate that deep prompt tuning is a very effective method. Specifically, the Macro F1 scores of DPMN are $0.8384$, $0.9218$, and $0.8165$ on the OLID, SOLID, and AbuseAnalyzer datasets. We prove that the proposed DPMN achieves excellent results in detecting abuse language.

Our main contributions are as follows: 
\begin{itemize}
  
  \item We propose a novel \textit{Deep Prompt Multi-task Network}, which achieves state-of-the-art results in detecting abuse language.
  \item Deep prompt tuning is first applied to abuse language detection. To improve the metric of detecting abuse language, we compare the forms of deep prompt tuning and light prompt tuning. Then the effects of different prompt lengths, tuning strategies, and prompt initialization methods are studied.
  \item We present a task head based on Bi-LSTM and FFN, and we prove that the effect of the task head is very significant through experiments.
  
\end{itemize}

\section{Related Works}\label{rewo}
\subsection{Abuse Language Detection}
Much past work has explored the problem of detecting different types of abuse language. On the whole, the detection of abuse language mainly includes the following types, aggression detection \cite{kumar:19}, bullying recognition \cite{huang:25}, hate speech recognition \cite{park:30}, abuse language, and toxic comments.
\begin{itemize}
\item Aggression detection: The developer can access a dataset of 15,000 tagged Facebook short texts as part of the shared challenge on aggression identification \cite{kumar:19} in TRAC-2018. It is applied to the model training and validation. In the performance test portion of the model, there are two distinct datasets used, one from Twitter and the other from Facebook. The detection task aims to distinguish three categories: non-aggressive, covertly aggressive, and overtly aggressive.
\item Bullying recognition: There are currently several works on cyberbullying detection methods. For example, Jun-Ming Xu \cite{xu:41} uses text classification, role labeling, sentiment recognition, and LDA to recognize related topics.
\item Hate speech recognition: Hate speech detection tasks have a long history of research \cite{kwok:43,burnap:44,djuric:45}. Davidson \cite{davidson:46} proposes a dataset for detecting hate speech that includes more than 24,000 tweets in English. 
\item offensive language: LSF \cite{chen:20} applies concepts from the theory of natural language processing to exploit the linguistic syntactic representations of sentences to detect offensive language. 
Zampieri \cite{Zampieri:33} introduces an offensive language recognition dataset OLID, which seeks to identify the class and the objective of offensive content in social networks. Rosenthal \cite{rosenthal:35} extend the OLID into the multilingual edition SOLID, which promotes multilingual research in detecting abuse language. MTL \cite{dai:17} uses multi-task learning and the BERT-base model to detect offensive language. 
\item Toxic comments: On Kaggle, there is a free contest called the Toxic Comment Classification Challenge. It provides the developer with short comments from Wikipedia. The dataset is divided into six groups: insult, obscene, threat, toxic, severe toxic, and identity hate. Through thorough trials on prompt engineering, Generative Cls \cite{wang:63} investigates the generative variation of zero-shot prompt-based toxicity detection. 

\end{itemize}
Although each task involves specific types of abuse or offense, many things are in common. For instance, insults against individuals are often called cyberbullying, and insults against groups are called hate speech.

MTL is the baseline model of our network. Compared with the MTL model, we first add prompt-based learning. Secondly, the task head is optimized. Our structure is Bi-LSTM + FFN, which is simpler and more effective than the LSTM + FFN + Softmax of MTL.

In contrast to Generative Cls, DPMN does not need to design manual prompts. Making a good prompt is very time-consuming and tough. In general, artificial design is not an intelligent solution.

\subsection{Prompt-based Learning for PLMs} 
To improve the output embedding from the PLMs, prompt-based learning entails adding instructions to the input text. With the development of GPT-3 \cite{brown:1}, prompt-based techniques have excelled in many common natural language processing applications. Many researches \cite{ben:2,lester:3,liu:4,lu:5,reynolds:6,scao:7} have been put forth to show how prompt-based learning has advanced by the appropriate manual prompt. Knowledgeable prompt-tuning \cite{hu:8} suggests calibrating the verbalizer to accommodate outside knowledge. Automatic generating for discrete prompt has been thoroughly investigated as a way to prevent time-consuming prompt design. LM-BFF \cite{gao:9} first explores the creation of label words and templates automatically. Additionally, Autoprompt \cite{shin:11} suggests using gradient-guided search to create the vocabulary template and identify terms automatically. Continuous prompts have recently been proposed\cite{hambardzumyan:12,he:13,li:14,liu:15}, which emphasize the use of learnable continuous representations rather than label words as prompt templates. In a word, prompt-based learning is applied in natural language processing to improve the understanding and generation of PLMs.

In view of the rapid rise and development of the above prompt tuning, we first design two continuous prompt forms, namely, deep prompt tuning and light prompt tuning. Then we apply the two prompt forms to abuse language detection. Eventually, we propose a network termed DPMN, which combines prompt-based learning and multi-task learning.

\section{The Proposed Methodology} \label{met}

\begin{figure*}[htp]
  \centering
  \includegraphics[width=12cm]{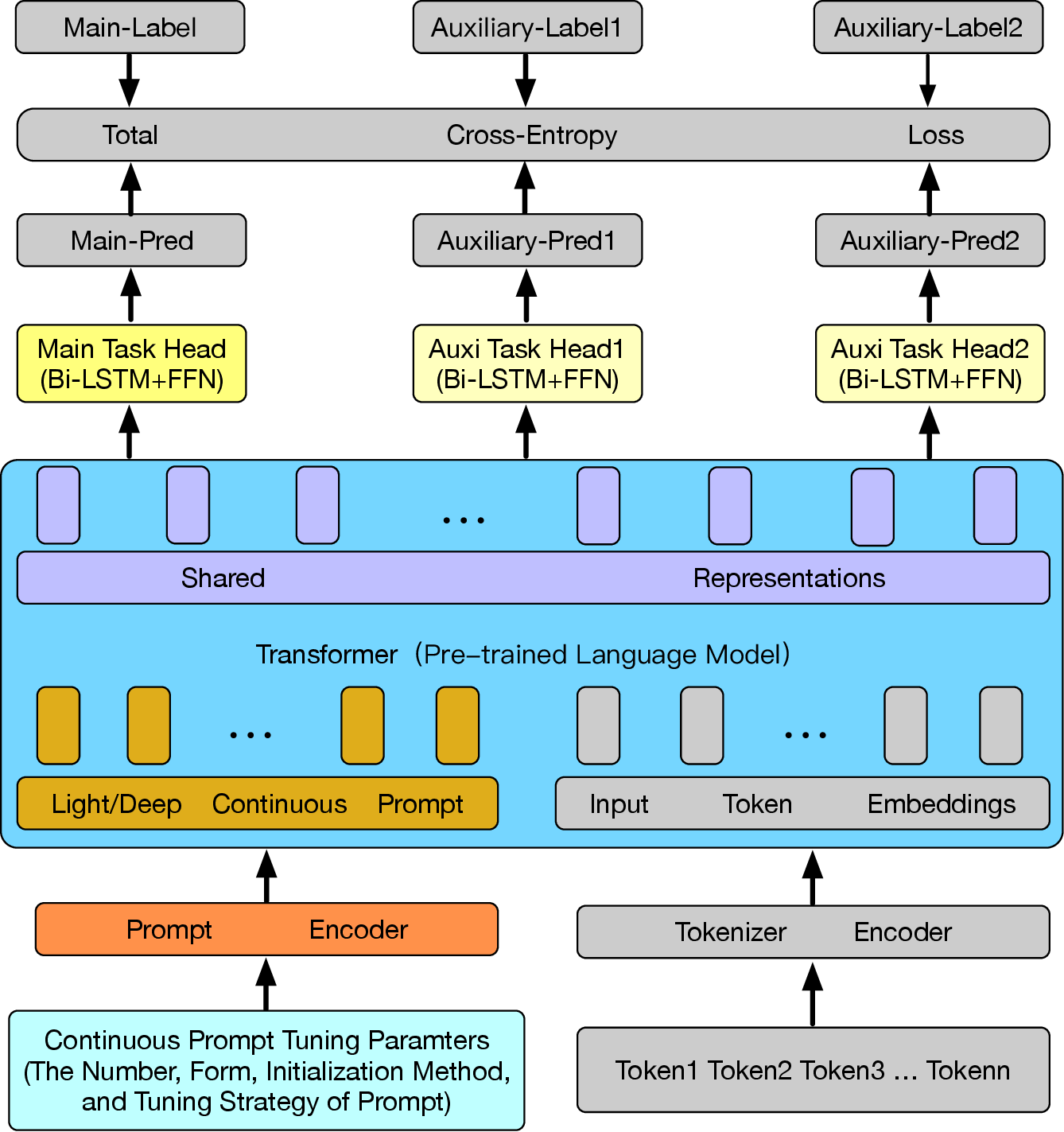}
  \caption{Our DPMN Architecture. According to the number of continuous prompt tokens, initialization strategy of continuous prompt, prompt form, and tuning strategy, the prompt encoder module generates the learnable embedding. The tokenizer encoder module encodes the short text to generate input embedding. We produce a representation matrix by combining the learnable embedding and input embedding. The representation matrix is input into PLMs. It outputs the shared embedding. The shared embedding is input to the task heads. The task heads output the probability value of the prediction classification. The total loss function is calculated to train the entire DPMN architecture.}
  \label{fig:01}
\end{figure*}

Our DPMN architecture is shown in Fig. \ref{fig:01}. The three sub-tasks share PLMs in the section at the bottom. Each sub-task has its unique module in the upper parts. A task head based on the Bi-LSTM and FFN neural network topology is present in each module.
DPMN sets the number, form, and initialization strategy of continuous prompt tokens. Then it generates the learnable embedding through the prompt encoder module. The tokenizer encoder module encodes the short text to generate input embedding. Splicing it with the input embedding generated by the PLMs. DPMN inputs them into the PLMs. The task head obtains the shared representations produced by the PLMs and generates a predicted category for short text. We calculate the multi-task loss function and train the entire architecture.

\subsection{Continuous Prompt Tuning Parameters}
Prompts can be divided into Discrete Prompts and Continuous Prompts.
\begin{itemize}
\item \textbf{Discrete Prompts.} Discrete prompts are essentially a natural language. Based on fixed prompt word templates, that is, adding fixed prompt word templates and masking words after inputting statements.

\item \textbf{Continuous Prompts.} Continuous prompts are no longer natural language. It replaces the fixed prompt word template with $n$ learnable vectors.
\end{itemize}

Continuous prompt tuning parameters are made up of the number, form, initialization method, and tuning strategy of prompt tokens. 

The number of prompt tokens plays a critical role in DPMN. We verify that abuse language detection usually achieves different performances with different prompt lengths. Specifically, we choose the number of prompt tokens according to the metric of detecting abuse language.

\begin{figure}[htp]
  \centering
  \includegraphics[width=7cm]{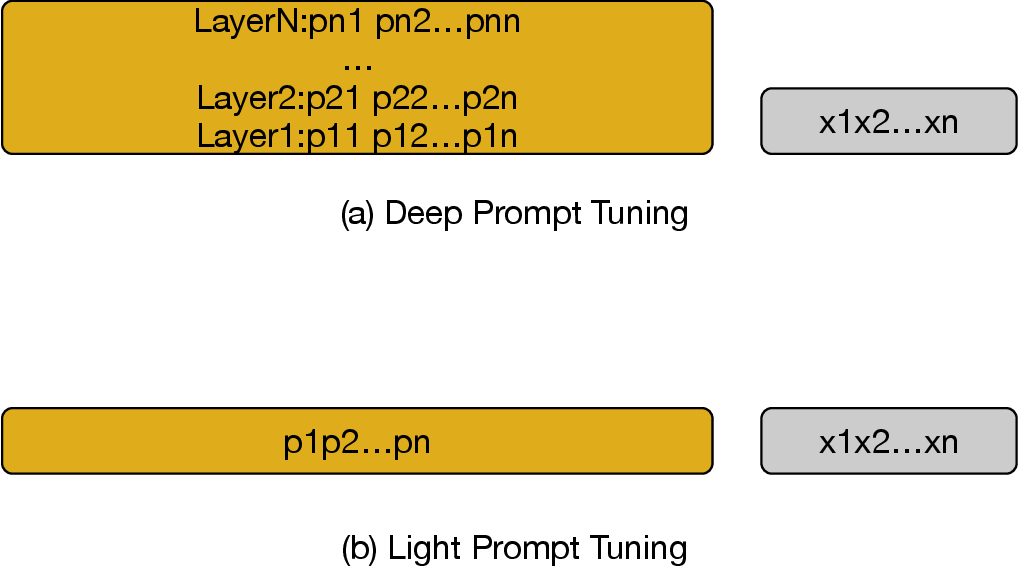}
  \caption{We design two continuous prompt forms, namely deep prompt tuning and light prompt tuning. The deep prompt tuning is to add trainable continuous prompt embedding to each layer of the PLMs. The light prompt tuning is to add trainable continuous prompt embedding to the first layer of the PLMs.}
  \label{fig:02}
\end{figure}

Considering the instability of discrete prompt performance, we adopt the form of continuous prompt tuning. Because the template of a discrete prompt requires a lot of manual design work, the predicted performance of the prompt-based model is relatively volatile. As shown in Fig. \ref{fig:02}, we use two continuous prompt forms: deep prompt tuning and light prompt tuning. We prove the effectiveness of these two prompt forms through experiments. The deep prompt tuning is to add trainable continuous prompt embedding to each layer of the PLMs. The light prompt tuning is to add trainable continuous prompt embedding to the first layer of the PLMs. For two continuous prompt forms, we study the most suitable prompt form.

In the architecture of our network, prompt initiation is a significant research challenge. It has two kinds of parameter initialization methods, which are random parameter initialization and BERT token initialization.

Tuning strategy is also an important research issue in our network design. It contains the two strategies of the Fixed LM Prompt Tuning strategy and LM + Prompt Tuning strategy.

\subsection{Prompt Encoder}
According to the number, form, initialization method, and tuning strategy of the prompt, the prompt encoder module generates a continuous prompt. Specifically, first define $n$ trainable embedding vectors, and then initialize them. Secondly, the input text is generated into a word ID sequence through a word splitter, and then an embedding sequence is generated based on the BERT model vocabulary. Finally, we concatenate these two vectors as inputs to the BERT model.

The PLMs are used as the backbone for our DPMN. The continuous prompt is applied to the encoder of the PLMs. To prepend continuous prompts for the encoder, DPMN initializes a trainable prefix matrix $p_{e}$ of dimension $p_{n} \cdot d$ for each layer of the PLMs, where $p_{n}$ is the prompt length, and $d$ is the hidden size of the PLMs. Because the prompt is located in the deep layers of the PLMs, it has the total capacity to guide the PLMs in the right direction and output a semantic shared representation for abuse language detection.

The continuous prompt stimulates the general knowledge of the PLMs. It performs better than fine-tuning in a range of natural language processing applications. The learnable embeddings are trained for abuse language detection.

\subsection{Task Head}
The central for prompt tuning is that the PLMs use a head to predict verbalizers. Because this requires manual design and even searches for all possible words, which is very labor-intensive. Therefore, the DPMN directly replaces the design of the verbalizer module with the classified label.
The task head of the DPMN adopts the neural network architecture of Bi-LSTM + FFN. Compared to the linear classification head, its predicted performance is better.
LSTM comprises an input gate, forget gate, output gate, and cell state. Bi-LSTM contained two sub-networks to model a text sequence in both directions. The outputs of short text are integrated in the following way:
\begin{equation}
H_{F,B}=\left[F_{-} H_t, B_{-} H_t\right],
\end{equation}
where $F_{-} H_t$ is output value of the LSTM at the last time $t$ in forward direction. $B_{-} H_t$ represents the output value of the LSTM at the last time $t$ in the backward direction. Here $H_{F, B}$ is the output embedding of the Bi-LSTM result.

FFN is made up of two linear layers. Its activation function is ReLU.
\begin{equation}
  y = W_{f2} ReLU(W_{f1}  H_{F,B} +b_{f1})+b_{f2},
\end{equation}
we let $\Theta_{FFN} = \{W_{f1}, W_{f2}, b_{f1}, b_{f2}\}$, where $\Theta_{FFN}$ is the learnable FFN parameter.
The task head takes the feed-forward network as the classifier. Multi-task classification is achieved by setting different output numbers of PLMs.

\subsection{Multi-task Network}
The proposed DPMN is a multi-task network, which is divided into the main task and two auxiliary tasks. Our main task is to detect whether the text is abuse language. The auxiliary tasks are to improve the feature representation ability of the output layer of the PLMs, thereby improving the detection ability of the main task and playing the role of transfer learning.

We train our DPMN on train datasets and mainly verify the model performance metrics on sub-task A. The goal of multi-task learning is to deliver useful information in tasks B and C to boost task A. 

\subsection{DPMN Loss}

\begin{equation}
  Loss_{sub-task}=-\sum_{i} y_{i} \log \left(y_{i}^{\prime}\right)
\end{equation}

$y_{i}^{\prime}$ is the probability predicted by the proposed DPMN.
$y_{i}$ is the category information of the dataset.
${Loss}_{sub-task}$ represents the loss of the sub-task, using the cross-entropy loss function.

\begin{equation}
  c_{main}+c_{auxi1}+c_{auxi2}=1
\end{equation}
$c_{main}$ is the loss coefficient for the main task. 
$c_{auxi1}$ is the loss coefficient for the auxiliary task. 
We set the sum of the coefficients of all current sub-task losses to 1.

\begin{equation}\label{loss}
  \begin{aligned}
    Loss_{\text{total}}=c_{main}Loss_{main}+c_{auxi1}Loss_{auxi1 }\\
    +c_{auxi2}Loss_{auxi2}
  \end{aligned}
\end{equation}

${Loss}_ {total}$ is the total loss of the DPMN. It is equal to the weighted sum of the losses of each sub-task.



\section{Experiments} \label{exp}
\begin{table*}[]
  \setlength{\tabcolsep}{3pt}
  \centering
  \caption{Four short tweets from the OLID, their corresponding labels are hierarchical. In task A, the aim is to discriminate between offensive and non-offensive posts. In task B, the goal is to predict the type of offense: Targeted Insult (TIN) and Untargeted (UNT). Task C focuses on the target of offenses: Individual (IND), Group (GRP), and Other (OTH).}
  \resizebox{\textwidth}{!}{
  \begin{tabular}{lllll}
    \toprule
    Short Text           &Task A   &Task B   & Task C   &  \\\midrule
    @USER With his offers, he is extremely kind.      & NOT & —   & —   \\
    Liberated! THE WORST ACTIVITY OF MY FUCKING LIFE & OFF & UNT & —     \\
    @USER This big cocksucker is fucked                  & OFF & TIN & IND   \\
    @USER Figures! Why are these people such idiots? Praise God for @USER & OFF & TIN & GRP  \\
    \bottomrule
  \end{tabular}}
  \label{Tab:00}
\end{table*}
We assess the proposed DPMN for detecting abuse language in experiments. Three public datasets are adopted: the OLID \cite{zampieri:34}, SOLID \cite{rosenthal:35}, and AbuseAnalyzer dataset \cite{chandra:16}. These datasets have been widely used for evaluating detection metrics of abuse language. Macro F1 score is used as the evaluation metric. DPMN achieves excellent performance in abuse language detection.

\subsection{Baseline}
To evaluate the detection metric, the proposed network DPMN is compared with eight comparable supervised methods, containing four shallow supervised methods (e.g., Logistic Regression \cite{kleinbaum:61}, XGBoost \cite{chen:58}, Bagging \cite{breiman:59}, SVM \cite{hearst:60}) and four deep supervised methods (e.g., BERT + Linear Head \cite{chandra:16}, MTL\cite{dai:17}, NULI \cite{liu:26}, UHH-LT \cite{wiedemann:55}).

\begin{table*}[!t] 
  \setlength{\tabcolsep}{4pt}
  \centering
  \caption{Detecting abuse language results on three public datasets, the evaluation metric is the Macro F1 score.}
  \resizebox{\textwidth}{!}{
  \begin{tabular}{llllllll}
    \toprule
    
    \textbf{Dataset}     & \textbf{Model}  & \textbf{Main Task} &  \textbf{Tuning Strategy}  & \textbf{Prompt Length} & \textbf{Prompt Initialization}\\  \midrule
    \multirow{6}{*}{OLID}  & Logistic Regression  & 0.7501   & *  & *    & *\\
    & Bagging  & 0.7558   & *  & *    & *\\
    & MTL & 0.8244   & LM Tuning  & *    & *\\
    & NULI  & 0.8290   & LM Tuning      & *  & *\\
    & DPMN(light prompt)   & 0.8279 & LM + Prompt Tuning & 1& Random\\
    & \textbf{DPMN(deep prompt)}          & \textbf{0.8384}   & \textbf{LM + Prompt Tuning} & \textbf{1} & \textbf{BERT Token}\\  \midrule
    \multirow{5}{*}{SOLID}  & MTL & 0.9139 & LM Tuning        & *  & * \\
    & MTL(Ensemble) & 0.9151 & LM Tuning        & *  & * \\
    & UHH-LT & 0.9204 & LM Tuning        & *  & * \\
    & DPMN(light prompt)   & 0.9208 & LM + Prompt Tuning & 1& Random\\
    & \textbf{DPMN(deep prompt)}          & \textbf{0.9218}  & \textbf{LM + Prompt Tuning} & \textbf{2} & \textbf{BERT Token} \\ \midrule
    \multirow{6}{*}{AbuseAnalyzer} & SVM  & 0.7277   & *         & * & *  \\
    & XGBoost  & 0.7157   & *         & * & *  \\
    & Logistic Regression  & 0.7235   & *         & * & *  \\
    & BERT + Linear Head & 0.7985   & LM Tuning          & * & *  \\
    & DPMN(light prompt)   & 0.8107 & LM + Prompt Tuning & 1& BERT Token\\
    & \textbf{DPMN(deep prompt)} & \textbf{0.8165} & \textbf{LM + Prompt Tuning} & \textbf{1} & \textbf{Random}  \\ 
    
    \bottomrule
  \end{tabular}}
  
  \label{Tab:01}
\end{table*}

\subsection{Analysis of Experimental Results}
The results in Table \ref{Tab:01} show detection metrics. The proposed DPMN is overall better than all the compared methods in three public datasets. For example, compared with the current state-of-the-art method, the Macro F1 score of our DPMN has been increased by 0.94\%, 0.14\%, and 1.80\% on the OLID, SOLID, and AbuseAnalyzer, respectively. 

The main reasons for these superior results come from three aspects:
\begin{itemize}
\item Deep prompt tuning can better use the general knowledge of the PLMs.
\item We propose an effective task head based on Bi-LSTM and FFN, and it improves the detection of abuse language.
\item DPMN utilizes multi-task learning, which can obtain more useful information from the other tasks. 
\end{itemize}
In addition, it can be seen that the detection effect of the deep models is better than that of the shallow models.

\begin{figure}[htp]
  \centering
  \includegraphics[width=8cm]{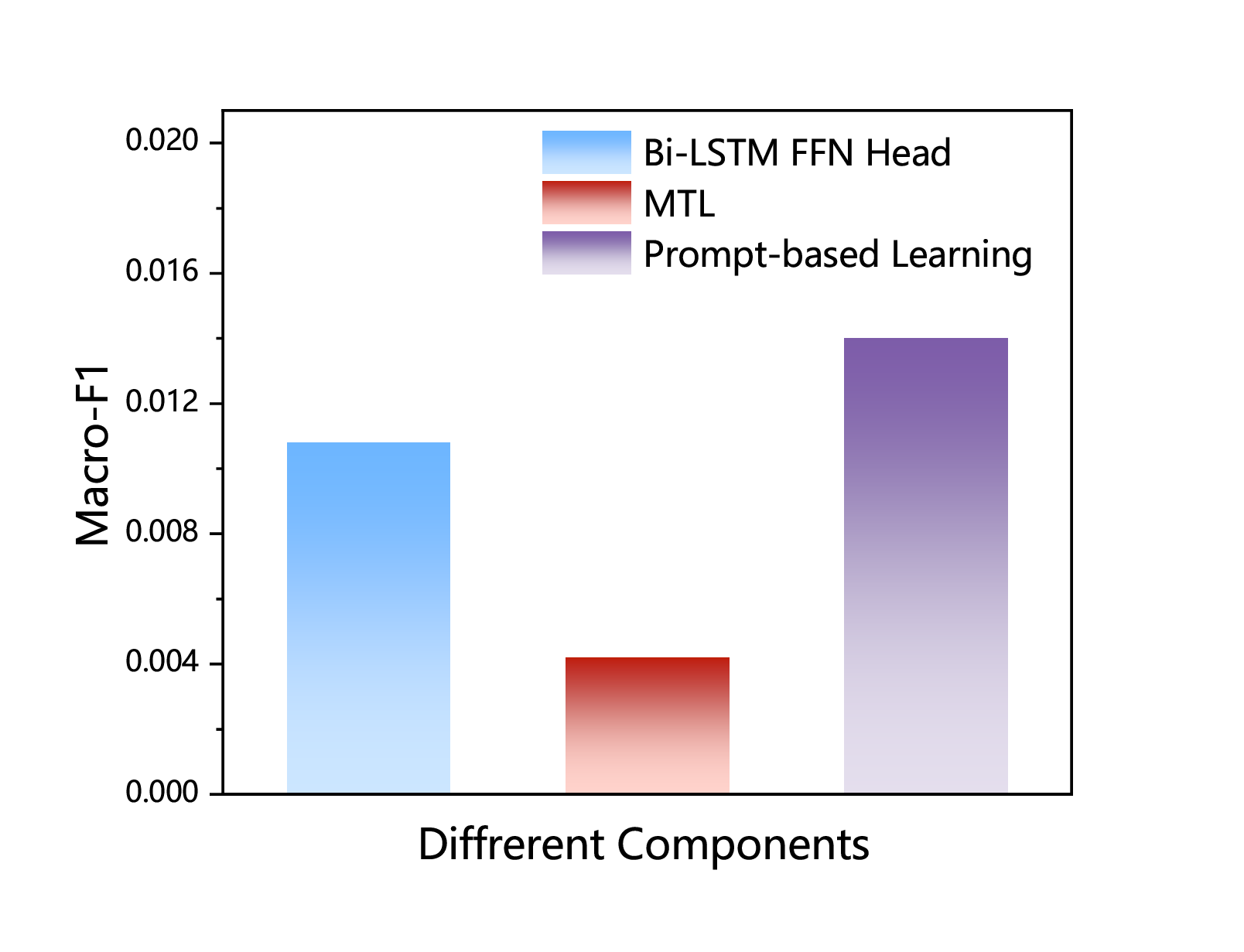}
  \caption{Through ablation experiments on the OLID dataset, the Macro F1 score contributions of different neural network modules of the DPMN algorithm are determined. A histogram of the structural lifting value is drawn in the network, which contains the Bi-LSTM + FFN module, the multi-task learning (MTL) module, and the prompt-based learning module.}
  \label{fig:03}
\end{figure}
\subsubsection{Ablation Experiment}

\begin{table*}[!t]
\centering
\caption{The Ablation Experiments of DPMN Components.}
\renewcommand{\arraystretch}{0.4}
\setlength{\tabcolsep}{3pt}
\resizebox{\textwidth}{!}{
\begin{tabular}{lp{10cm}p{2cm}} \toprule
\textbf{Model}    & \textbf{Architecture}                   & \textbf{Macro F1}   \\ \midrule
BERT Base & BERT + Linear Head                          &0.8037  \\
BERT LSTM & BERT + Bi-LSTM FFN Head                         & 0.8202     \\
DPMN-P       & BERT + Bi-LSTM FFN Head + Multi-task Learning   & 0.8244    \\
DPMN-M       & BERT + Bi-LSTM FFN Head + Prompt   & 0.8342    \\
DPMN-B       & BERT + Multi-task Learning + Prompt   & 0.8276    \\
DPMN      & BERT + Bi-LSTM FFN Head + Multi-task Learning + Prompt& 0.8384    \\ \bottomrule
\end{tabular}}
  
\label{Tab:02}
\end{table*}

The results in Table \ref{Tab:02} show the ablation experiment. We design the ablation experiment for the DPMN components. \textit{BERT Base} model adopts the BERT + Linear Head structure, where the Linear Head is a classifier of multi-layer perceptron structure. The Macro F1 score of the \textit{BERT Base} model to detect abuse language is $0.8037$.  The \textit{BERT LSTM} model utilizes the BERT + Bi-LSTM FFN Head structure, where the Bi-LSTM FFN Head is a classifier of the Bi-LSTM + FFN structure. The Macro F1 score of the \textit{BERT LSTM} model to detect abuse language is $0.8202$. We experimentally prove that the classification head based on Bi-LSTM + FFN is better than the classification head based on Linear Head.

\textit{DPMN-P} adds a multi-task learning architecture based on the \textit{BERT LSTM} model. It removes the prompt tuning module compared to DPMN. The Macro F1 score of the \textit{DPMN-P} model to detect abuse language is $0.8244$. DPMN optimizes the \textit{DPMN-P} model and designs the architecture of prompt-based learning. The Macro F1 score of the DPMN model to detect abuse language is $0.8384$. Comparing the detection performance of \textit{DPMN-P} and DPMN models, we can conclude that deep continuous prompt learning is effective.

In contrast to DPMN, \textit{DPMN-M} gets rid of the multi-task learning module. Its Macro F1 score is $0.8342$.

Compared to DPMN, \textit{DPMN-B} does away with the Bi-LSTM + FFN Head. Its metric value is $0.8276$.

To express the validity of the ablation experiment, we draw a histogram of the structural lifting value in the network. From the above Fig. \ref{fig:03}, in the evaluation value of Macro F1, the contribution of the Bi-LSTM + FFN module is $0.0108$, the contribution of the multi-task learning (MTL) module is $0.0042$, and the contribution of the prompt-based learning module is $0.0140$. Therefore, it can be seen that the proposed DPMN is effective.



\subsubsection{The Convergence of DPMN}
\begin{figure}[htp]
  \centering
  \includegraphics[width=8cm]{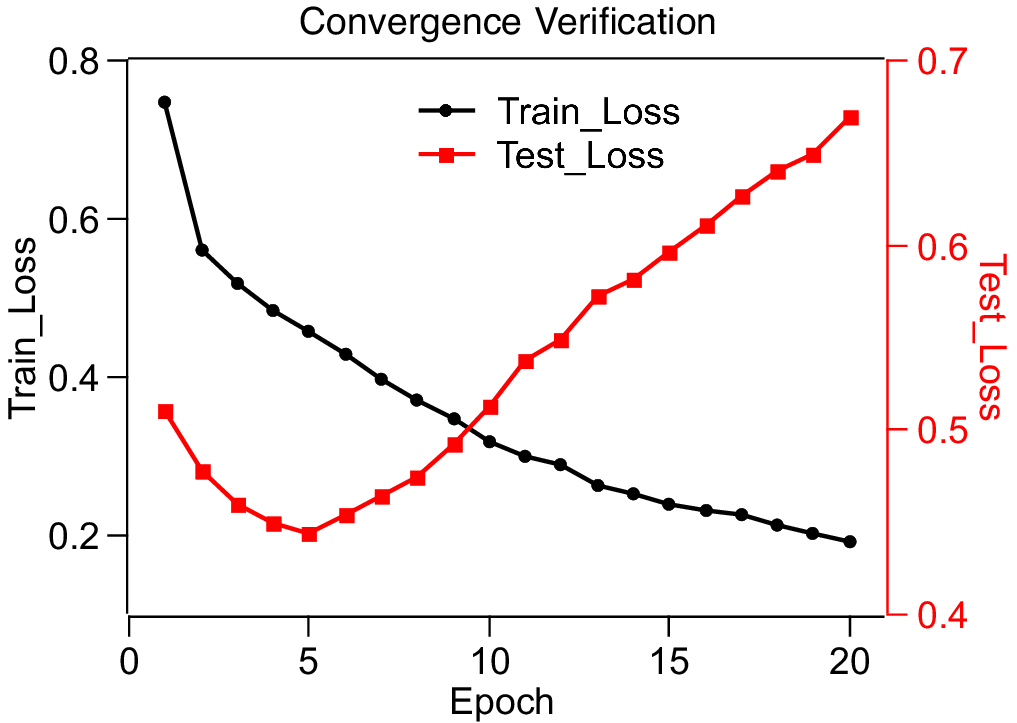}
  \caption{The convergence of the DPMN is verified on the OLID dataset. }
  \label{fig:06}
\end{figure}

From Fig. \ref{fig:06}, the training loss curve shows a downward trend with the increase in the number of epochs. As the number of epochs increases, the test loss curve first decreases and then increases. With the deepening of DPMN training, the test loss shows an upward trend, indicating that the detection performance of DPMN is declining. Through the changes in the two curves, it can be concluded that the DPMN has the best performance at $epoch=5$.

\subsection{Implementation Details}
We chose a learning rate of $3e-6$ and a batch size of $32$ for our best DPMN. The loss coefficients for sub-tasks A, B, and C are $0.4$, $0.3$, and $0.3$, respectively. We use an early stop method to stop tuning the model if the validation Macro F1 does not rise in four consecutive epochs. We train the DPMN with a maximum of $30$ epochs. The DPMN is implemented in PyTorch, and a single GPU-V100 is used for each experiment.

\section{Conclusion and Future Work} \label{cofu}
We propose an innovative \textit{Deep Prompt Multi-task Network} termed DPMN. It introduces deep prompt tuning in abuse language detection for the first time. It can better motivate the knowledge of PLMs. We design a task head based on Bi-LSTM and FFN, which improves the performance in detecting abuse language. We attempt two prompt forms and verify the effects of different prompt lengths, tuning strategies, and prompt initialization methods. The proposed DPMN achieves state-of-the-art results in three abuse datasets. The follow-up work is to optimize and adaptively adjust the sub-task loss weight in multi-tasks and reasonably design the algorithm of the whole model loss function.

\section{Acknowledgment}
This work was supported by the National Key Research and Development Program of China (Grant No. 2021ZD0201501), the Youth Foundation Project of Zhejiang Province (Grant No. LQ22F020035), and the National Natural Science Foundation of China (No. 32200860).

\bibliography{main}

\end{document}